\documentclass[10pt]{article}
\usepackage[english]{babel}
\usepackage{multirow} 
\usepackage{graphicx}
\usepackage{amsmath}
\usepackage[utf8]{inputenc}  
\usepackage{booktabs}   
\usepackage{authblk}
\usepackage{mystyle}
\usepackage[numbers,sort&compress]{natbib}
\usepackage{xcolor}
\definecolor{darkgreen}{rgb}{0,.5,0}
\usepackage[colorlinks,filecolor=blue,citecolor=darkgreen,unicode]{hyperref}
\title{Optimizing IoT Threat Detection with Kolmogorov-Arnold Networks (KANs)}
\author{Natalia Emelianova\thanks{email: natalia.emelianova@ufabc.edu.br},\ Carlos Kamienski\thanks{email: carlos.kamienski@ufabc.edu.br}\\ and \\ Ronaldo C. Prati\thanks{email: ronaldo.prati@ufabc.edu.br}}
\affil{\small Federal University of ABC (UFABC)\\ Avenida dos Estados, 5001, Bang\'u, \\CEP 09280-560 -- Santo Andr\'{e} -- SP -- Brazil\\ Center for Mathematics, Computing and Cognition (CMCC)}
\date{} 
\begin{document} 
	
	\maketitle
	
	\begin{abstract}
		The exponential growth of the Internet of Things (IoT) has led to the emergence of substantial security concerns, with IoT networks becoming the primary target for cyberattacks. This study examines the potential of Kolmogorov-Arnold Networks (KANs) as an alternative to conventional machine learning models for intrusion detection in IoT networks. The study demonstrates that KANs, which employ learnable activation functions, outperform traditional MLPs and achieve competitive accuracy compared to state-of-the-art models such as Random Forest and XGBoost, while offering superior interpretability for intrusion detection in IoT networks.
		
		\textbf{Keywords.} Internet of Things (IoT), Intrusion Detection System (IDS), Kolmogorov-Arnold Networks (KANs), Machine Learning, Feature Selection, IoT Security, Network Traffic Classification.
	\end{abstract}

	\section{Introduction}
	The Internet of Things (IoT) is a rapidly expanding network of interconnected devices that enable smart functionality in various applications, from healthcare and industrial automation to smart cities and home systems~\cite{Atzori2017}. Although IoT offers significant advantages in terms of automation, efficiency, and convenience, it also introduces substantial security challenges. As IoT devices become more integral to critical infrastructure, they are increasingly targeted by malicious actors, exposing sensitive data and systems to cyberattacks. As a result, manufacturers and academics now have the top priority of improving the security of IoT devices.  In recent years, significant efforts have been made to address security concerns in the IoT paradigm~\cite{Sarker2022,Sasi2024}. Traditional Intrusion Detection Systems (IDS) often struggle to keep up with the scale and complexity of IoT environments. These systems typically rely on fixed feature sets and predefined patterns to detect threats, making them ill-suited to handle the dynamic and evolving nature of IoT attacks~\cite{AbdElaziz2024}. 
	
\begin{figure}[ht]
		\centering
		\includegraphics[width=0.9\linewidth]{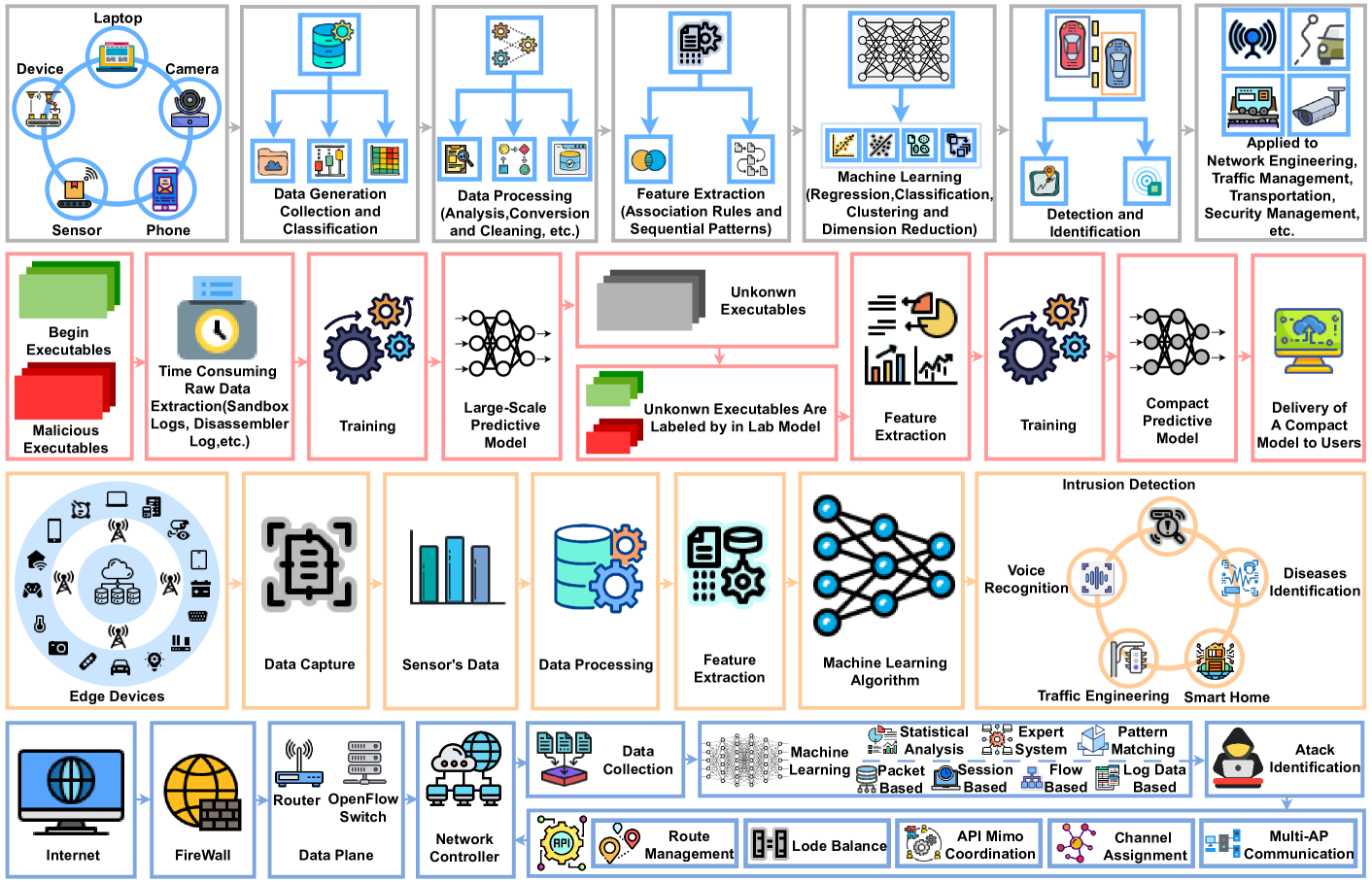}
		\caption{Typical scenarios and ML-based IoT visions \cite{Liu2024}.}
		\label{fig1}
	\end{figure}

	One of the most effective methods to enhance the intelligence of data analysis and processing in the Internet of Things (IoT) is to integrate Machine Learning (ML) into its operations~\cite{Liu2024}. ML systems, as shown in Fig.\,\ref{fig1}, are designed to automatically learn models from training data and use these models to make predictions. They find extensive applications across numerous fields, including ML-based IoT device identification, ML-based IoT Malware, etc.
	ML serves as a crucial technology, not only for analyzing IoT-generated data, but also for diverse applications within the IoT ecosystem. For example, ML finds utility in IoT device recognition, anomaly detection, and even in uncovering malicious activities. IoT attacks are cyberattacks that use any IoT device to access sensitive consumer data. As IoT devices are not projected with adequate safety mechanisms, they are one of the weakest links in an organization and offer a significant risk of security~\cite{waqaskhan2024, kaur2023}.
	
	The study of existing methods to find attacks on IoT shows that scientists and experts in the area are actively involved in solving the vulnerability in the field of security. However, some obstacles in the field of threat detection strategies require the development of more innovative approaches and additional research \cite{Neto2023,Neto2024,Cvitic2022}.
	
	This work explores the use of Kolmogorov-Arnold networks (KAN) \cite{https://doi.org/10.48550/arxiv.2404.19756}, a novel machine learning architecture inspired by the Kolmogorov-Arnold representation theorem, to improve intrusion detection in IoT networks. Unlike MLPs with fixed activation functions, KANs use spline-based learnable activations, enabling dynamic adaptation to complex data patterns, a critical advantage for evolving IoT threat landscapes. Using feature selection and KANs, this research aims to optimize both detection performance and computational efficiency in real-time IoT environments.
	
	Due to the fact that accurate detection of anomalies is crucial in IDS, and KANs offers a promising alternative to MLP-based approaches, this paper proposed using the rebuilding of the CNN model by replacing all MLP layers with KAN layers~\cite{AbdElaziz2024}. The ability of KANs to approximate complex functions aligns well with the needs of IDS in IoT environments, where computational limitations are critical considerations~\cite{Elsaid2024}.
	The primary objectives of this study can be listed as follows:
	
	\begin{itemize}
		\item Evaluate the performance of various machine learning models (Logistic Regression, Random Forest, Decision Trees, K-Nearest Neighbors, Gradient Boosting, XGBoost, Naive Bayes, Multi-Layer Perceptron, AdaBoost) in terms of precision, recall, F1-score, training time, and prediction time for IoT intrusion detection.
		
		\item  Implement and evaluate the KAN model, focusing on its ability to detect intrusions in IoT networks, particularly through its use of learnable activation functions and feature selection.
		
		\item 	Analyze the impact of feature selection on model performance, particularly in relation to reducing computational overhead without sacrificing accuracy.
		
		\item Investigate the suitability of KANs for real-time IoT applications, evaluating both their accuracy and computational efficiency.
		
	\end{itemize}
	
	This research contributes to the growing body of knowledge in IoT security by demonstrating the effectiveness of KANs in intrusion detection tasks. Key contributions include:
	
	\begin{itemize}
		\item Application of KANs to the CIC IoT 2023 dataset \cite{Neto2023}, showing how learnable activation functions at the edges can improve the accuracy and interpretability of the model.
		
		\item Evaluation of KANs against traditional models (e.g., Random Forest, XGBoost) to demonstrate their competitive performance and suitability for IoT intrusion detection.
		
		\item Demonstration of KANs' interpretability through symbolic formula generation, enabling transparent decision-making in security-critical IoT systems.

	\end{itemize}
	
	This work opens new avenues for improving the scalability, interpretability, and real-time application of intrusion detection systems in IoT environments.
	
	The remainder of this paper is structured as follows: Section \ref{sec:rel_work} presents the background and related work from the literature, introduces the basics of the IoT architecture, outlines the types of IoT attacks that can be perpetrated, and provides an overview of Kolmogorov-Arnold Networks. Section \ref{sec:method} presents the steps of the proposed method. Section \ref{sec:experiments} presents the findings of the developed model and offers a critical analysis thereof. Finally, Section \ref{sec:discussion} presents a discussion of the findings, and Section \ref{sec:conclusion} offers conclusions and suggestions for future research.
	
	\section{Background and Related Work}\label{sec:rel_work}
	
	The application of machine learning to threat detection in the Internet of Things is a fast-growing research area. Despite significant progress, there are several areas in which more research and development is needed~\cite{Wang2024, Lazzarini2023, Tsimenidis2021, ArnauMuoz2024, MohyEddine2023}. With the development of IoT, new types of attack are emerging and those existing are constantly evolving. The emergence of new, more complex and adaptive attacks, difficult to detect by existing machine learning models, leads to the need to develop models that are capable of continuous learning and adaptation to new types of threats, as well as to use the methods and Generative models for creating synthetic training data~\cite{arifin2024surveyapplicationgenerativeadversarial, Lim2024}.
	
	There is a problem of lack of quality and quality data for training models, especially for new types of devices and attacks. The class imbalance (when the number of normal samples significantly exceeds the number of attacks) worsens the quality of the classification~\cite{Mahdavifar2024,Wang2021,ANOH2024}. Furthermore, the difficulty in interpreting solutions obtained from machine learning models, especially deep neural networks, often makes it difficult to understand the causes of false positives and lost attacks. The resolution of this problem requires the development of methods to view and explain model solutions \cite{Qaddos2024}.

	Safety problems and possible solutions for IoT systems are constantly being studied. Several research papers are dedicated to a comprehensive rating of attacks based on several factors. Real perspectives and perspectives of this research area are being taken into account \cite{Sasi2024}.
	
	Recent studies explored explainable AI (XAI) techniques in graph-based IDSs and neural-symbolic systems \cite{Mahdavifar2024,Kalutharage2024}. Ensemble models and deep learning approaches offer strong accuracy, but often lack interpretability \cite{Rane2024} -- a gap that KANs aim to bridge. KANs offer a symbolic, transparent framework that can be especially beneficial in high-stakes detection scenarios \cite{kilani2025}.
	
	\subsection{IoT architecture}
	
	The IoT is a network of smart assets deployed in various locations characterized by its openness and comprehensiveness. In recent years, significant efforts have been made to address security concerns within the IoT paradigm. Specific techniques in the realm of IoT security focus on addressing security concerns at a particular layer, while alternative approaches strive to ensure comprehensive end-to-end security for IoT systems \cite{Sasi2024}.
	
	The IoT architecture can be categorized into seven levels: perception, transport, edge, processing, application, business, and security layer. The system operates as a closed loop, facilitating the production of customized goods tailored to meet each end-customer's particular requirements. Rapid growth of the IoT requires the implementation of robust security and privacy protocols to mitigate potential system vulnerabilities and threats. In addition, within the IoT realm, factors such as dependability, scalability, and power consumption emerge as crucial considerations. In the present setting, conventional security measures may not always be suitable.
	
	The absence of standardized protocols in the IoT architecture poses more challenges regarding interoperability, security, and several other concerns. The IoT architecture has the potential to encompass a maximum of seven layers \cite{simmons2022internet}. The \textbf{\textit{perception layer}} of an IoT system architecture, also known as the device layer, consists of multiple elements: sensors, cameras, actuators, and similar devices that collect data and perform tasks. The \textbf{\textit{transport layer}} of an IoT system architecture transmits data from multiple devices (e.g., on-site sensors, cameras, actuators) to an on-premise or cloud data center. As IoT networks grow on a scale, to process and analyze data as close as possible to the source, the \textbf{\textit{edge layer}} of the architecture of the IoT Edge Computing system is used. A fundamental component of an IoT system architecture is its \textbf{\textit{processing layer}}, also called the middleware layer, which typically leverages many connected computers simultaneously, in the form of cloud computing, to deliver superior compute, storage, networking, and security performance. The \textbf{\textit{application layer}} of an IoT system architecture involves decoding promising patterns in IoT data and compiling them into summaries that are easy for humans to understand, such as graphs and tables.
	
	Patterns decoded at the application level can be used to further distill business insights, project future trends, and drive operational decisions that improve efficiency, safety, cost effectiveness, customer experience, and other important aspects of business functionality. In fact, all this can be achieved at the \textbf{\textit{business layer}} of an IoT system architecture. The IoT \textbf{\textit{security layer}} comprises three main aspects:  (1)\textit{Equipment Security} involves actual IoT devices and protects these endpoints from malware and hijacks; (2)\textit{Cloud Security} with most IoT data being processed in the cloud, cloud security is crucial to prevent data leaks; (3)\textit{Connection Security} focused on securing data transmitted across networks, primarily with encryption. The Transport Layer Security (TLS) protocol is considered the benchmark for the security of IoT connections.
	
	\subsection{IoT attacks}
	
	IoT attacks bring new problems that require specialized security solutions to fully guard against these dangers. Some of the different ways are \cite{Sasi2024}:
	
	\begin{description}
		
		\item [Attack Surface] IoT devices often have low processing speeds and resources.
		\item [Diversity of Devices] The types of IoT devices differ significantly in form factor, operating systems, and network connection.
		\item [Physical Impact] IoT devices are frequently used in crucial infrastructure or life-sustaining systems, such as medical equipment; therefore, a cyberattack on these devices might have very harmful physical consequences. 
		\item [Legacy devices] IoT devices usually have a longer lifespan. Older devices cannot get software updates or security patches, making them more vulnerable to attacks or compromises.
	\end{description}
	
	\subsection{Kolmogorov--Arnold Networks (KANs)}
	
	KANs are an advanced alternative to MLP \cite{https://doi.org/10.48550/arxiv.2404.19756}. The theoretical foundation of KANs is based on the Kolmogorov-Arnold representation theorem \cite{Kolmogorov:1957,Arnold:1957},  which shows that any continuous multivariate function can be represented as a superposition of univariate functions and addition. It seeks to overcome limitations in MLPs through a new architecture where each weight parameter is replaced by a learnable one-dimensional function, most often parametrized as a spline. It overcomes the rigidity of MLPs through flexibility and adaptability at the level of a single connection within the network \cite{https://doi.org/10.48550/arxiv.2404.19756}.
	
	Although MLPs use fixed activation functions at the nodes, they use a linear transformation across layers. KANs apply the learnable activation functions directly on the network edges. This, in turn, removes the complete requirement for linear weight matrices. Instead, the nodes in KANs add up the incoming signals to produce a result, making it less computationally expensive and enhancing the model for precise approximations of functions. KANs inspired by this mathematical framework \cite{https://doi.org/10.48550/arxiv.2404.19756}.
	\begin{equation}
		f\left ({{x_{1}, \ldots , x_{n}}}\right )=\sum _{q=1}^{2 n+1} \Phi _{q}\left ({{\sum _{p=1}^{n} \phi _{q, p}\left ({{x_{p}}}\right )}}\right ) \label{1}.
	\end{equation} 
	Here $\Phi _{q}$ and $\phi _{q, p}$ are univariate learnable functions that collectively handle the input-to-output mapping of the network. This setup allows KANs to handle high-dimensional data by breaking down the complexity into manageable, one-dimensional operations, thus sidestepping the curse of dimensionality that plagues MLPs. KANs use splines for the activation functions on network connections. 
	
\begin{figure}[tb]
		\centering
		\includegraphics[width=0.35\linewidth]{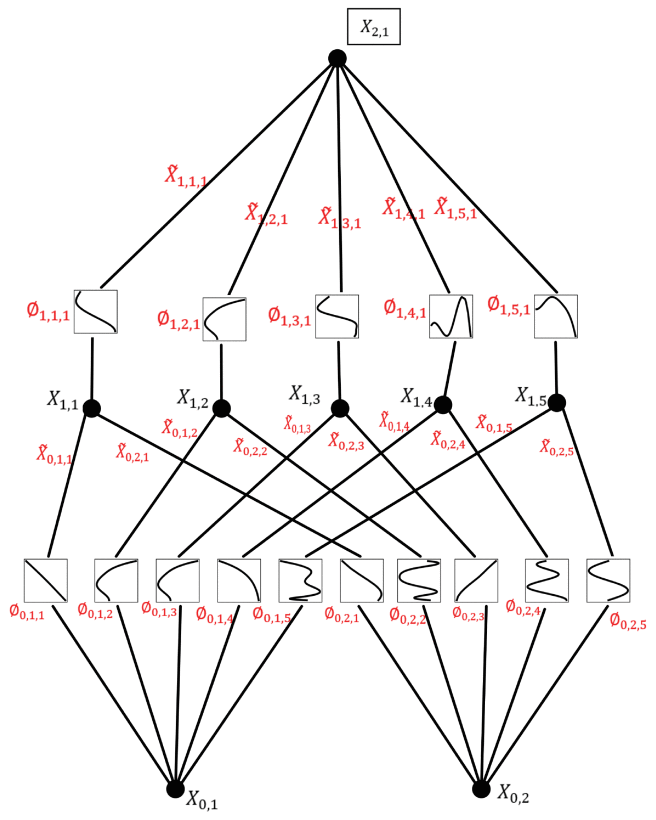}
		\caption{Activation function that flow via the network \cite{AbdElaziz2024}.}
		\label{fig3}
	\end{figure}

	\begin{description}	     
		
		\item[Spline Activation Functions:] Each connection in the network uses a spline function defined by the following formula \cite{https://doi.org/10.48550/arxiv.2404.19756}
		\begin{equation} 
			s(t) = \sum _{i=1}^{k} c_{i} B_{i}(t) \label{2},
		\end{equation} 
		where $ B_{i}(t)$ are the spline basis functions and $c_{i}$ are coefficients learned during training. Fig.\,\ref{fig3} depicts the splines function. It shows a network diagram with nodes connected by functions $\phi _{i, j}$  where each function processes input from the preceding nodes and the outputs are summed at the subsequent nodes.
		
		\item[Network Function Representation:] The network is described by the function \cite{https://doi.org/10.48550/arxiv.2404.19756}
		\begin{equation}
			f(x) = \sum _{j=1}^{m} a_{j} \sigma _{j}(w_{j}^{T} x + b_{j}),\label{3}
		\end{equation} 
		where $\sigma _{j} $ represent the spline-based activation functions, $w_{j}$ are the input weights, $b_{j}$ are biases, and $a_{j}$ are output weights.
	\end{description}

	Table~\ref{tab:model_comparison} presents a comparison of the main aspects of the KAN model with other ML algorithms. KANs uniquely combine symbolic interpretability and effective dimensionality handling through learnable splines, making them suitable in IoT contexts.

	\begin{table}[ht]
		\centering
		\caption{Comparison of Machine Learning Models for IoT Intrusion Detection}
		\label{tab:model_comparison}
		\resizebox{\textwidth}{!}{
			\begin{tabular}{@{}p{3cm}p{3.5cm}p{3.5cm}p{3.5cm}p{3.5cm}p{3.5cm}@{}}
				\toprule
				\textbf{Aspect} & \textbf{KANs} & \textbf{MLPs} & \textbf{Tree-Based Models} & \textbf{kNN} & \textbf{Logistic Regression} \\
				\midrule
				\textbf{Activation} & Learnable splines (edges) & Fixed functions (nodes) & N/A & N/A (distance-based) & Sigmoid (output layer only) \\
				\addlinespace
				\textbf{In\-ter\-pre\-ta\-bi\-li\-ty} & High (symbolic formulas) & Low (black-box) & Moderate (feature importance) & Moderate (instance-based) & High (coefficients) \\
				\addlinespace
				\textbf{Training Time} & High & Moderate & Low & Low (lazy learner) & Low  \\
				\addlinespace
				\textbf{Prediction Time} & Moderate & Moderate & Low & High & Low  \\
				\addlinespace
				
				\textbf{Curse of Dimensionality} & Mitigated via univariate functions & Struggles & Handles well with selection & Severe degradation & Moderate (with regularization) \\
				\bottomrule
			\end{tabular}%
		}
	\end{table}
	
	\section{Methodology}\label{sec:method}
	
	The first part of this study investigates the performance of different machine learning models to detect intrusions in IoT networks. We explored both classical and advanced techniques, considering not only their predictive accuracy, but also their computational efficiency in terms of training and prediction times, as well as  interpretability. 
	
	\subsection{Data selection and Preprocessing} \label{sec:method_1}
	
	The dataset used in this experiment is the CIC IoT 2023 Dataset, which contains both benign and malicious network traffic data.  This dataset is from the Center for Cybersecurity of the University of New Brunswick.  Presented in \cite{Neto2023}, the dataset supports the development and evaluation of intrusion detection systems. This dataset is very detailed and provides a broad and practical testbed to assess the effectiveness of security solutions tailored to the diverse range of IoT-specific cyber threats. It has extracted CSV features from network traffic in 105 IoT devices with 33 cyberattacks running on them. Seven types of attacks were present: distributed denial of service (DDoS), denial of service (DoS), reconnaissance, web-based, brute force, spoofing and Mirai botnet, as summarized in Table \ref{tab1}. 
	In our study, data binarization is applied  to facilitate the analysis of the classification task. The dataset contains a categorical target variable with two classes:  \texttt{"BenignTraffic"} (representing normal traffic) and  \texttt{"MaliciousTraffic"} (representing attack traffic).
	These labels are converted into binary form:
	\begin{equation}
		y =
		\begin{cases}
			1, & \text{if the sample belongs to \texttt{BenignTraffic}} \\
			0, & \text{if the sample belongs to \texttt{MaliciousTraffic}}
		\end{cases}
	\end{equation}
	This binarization process ensures that the data are properly formatted, allowing it to efficiently learn and distinguish between normal and malicious network traffic.
	
	\begin{table}[htbp]
		\caption{Description of the CICIoT2023 Dataset}
		\begin{center}
			\begin{tabular}{|c|c|}
				\hline
				\textbf{Property}&{\textbf{Value}} \\
				\hline
				Number of Classes&  2 (combination of all attack classes and the benign class)  \\
				\hline
				Number of Samples&  1 048 575  \\
				\hline
				Number of Features&  47 \\
				\hline
			\end{tabular}
			\label{tab1}
		\end{center}
	\end{table}
	
	The dataset is loaded using libraries such as Pandas, PyTorch and  Scikit-Learn \cite{reback2020pandas,mckinney-proc-scipy-2010}. The data were split into a training set (67\%) and a test set (33\%) using the $train\_test\_split$ function to consistently evaluate model performance. 
	
	\subsection{Models Evaluated} \label{sec:method_2}
	
	To compare results with KAN network, we evaluated the following machine learning algorithms: Logistic Regression, Random Forest, Decision Trees, K-Nearest Neighbors (KNN), Gradient Boosting, XGBoost, Naive Bayes, Multi-Layer Perceptron (MLP), and AdaBoost. 
	
	For each trained model, we generate a classification performance that includes the precision, recall, F1 score, and overall accuracy of the measures. These metrics provide a comprehensive insight into the model's capability to classify intrusions in IoT environments. We also measure the computational efficiency, in terms of training time and prediction time. For each machine learning model, the training and prediction processes are timed and followed by a performance report on the test set.

	\subsection{Building of KAN model} \label{sec:method_3}
	
	The second part of the research is devoted to building a model of the KAN \cite{Liu2024}. The network topology is defined as follows:
	\begin{itemize}
		\item Input Layer: the number of neurons is equal to the number of features selected ($\textit{input\_ dim}$).
		\item Hidden Layers: two hidden layers with 16 and 8 neurons, respectively, using the MultiKAN architecture to allow for both additive and multiplicative interactions between features. 
		\item Output Layer: a final layer with 2 neurons, representing the classification into "BenignTraffic" and "MaliciousTraffic".
	\end{itemize}
	
	Mathematically, the structure of the model can be represented as: $model=KAN(width=[input\_ dim,[16,8],2])$ where each functional unit within the network follows the Kolmogorov-Arnold representation, allowing for a more expressive function approximation compared to conventional neural networks.
	
	The features are first normalized using \textit{StandardScaler}, which transforms each feature to have zero mean and unit variance, regardless of whether feature selection is applied or not: \begin{equation} X_{\text{scaled}} = \frac{X - \mu}{\sigma}, \end{equation} where $\mu$ is the mean and $\sigma$ is the standard deviation of each feature in the dataset.
	
	Next, feature selection is applied by ranking features according to their importance and retaining only the top $N$ most relevant ones. The resulting scaled feature set is then transformed into PyTorch tensors for use in the KAN model: \begin{equation} X_{\text{train}}, X_{\text{val}}, X_{\text{test}} \in \mathbb{R}^{m \times N}, \end{equation} where $m$ is the number of samples and $N$ is the number of selected features.
	
	The corresponding binary class labels are also converted into \textit{long tensors}, as required for classification tasks.
	
	The dataset used in this study \cite{Neto2023} consists of $1\,048\,575$ samples, each with $47$ features (Tab.\ref{tab1}). For training and evaluation purposes, the dataset is split into three parts: training, validation, and test sets, using the \textit{train\_test\_split} function from the \textit{scikit-learn} library.
	
	The splitting is done in two stages.
	First, the dataset is divided into training data (70\%) and temporary data (30\%):
	\begin{itemize}
		\item Training set: 70\% $\Rightarrow$ 734,002 samples
		\item Temporary set: 30\% $\Rightarrow$ 314,573 samples
	\end{itemize}
	
	Next, the temporary set is equally split into validation and test sets (each 15\% of the total data):
	\begin{itemize}
		\item Validation set: 15\% $\Rightarrow$ 157,286 samples
		\item Test set: 15\% $\Rightarrow$ 157,287 samples
	\end{itemize}
	
	The model was initially trained using all the available features in the dataset. The $feature\_importances\_ attribute$ of the trained Random Forest model was then used to assess the importance of each feature. The features were ranked according to their importance and the top 10 features were selected for further analysis. A new training set was created by selecting only the top 10 features of the original training set. Similarly, the validation set was created by selecting the corresponding same top 10 features from the original validation set.
	
	The KAN model is trained using the training dataset, using Adam optimizer CrossEntropyLoss. The model undergoes multiple training iterations, in which it learns to minimize the loss function by adjusting its parameters based on the training data. For the whole learning process, the model performs 114680 iterations. The number of iterations for the entire learning process is:
	\begin{equation}
		I_{\text{total}} = \frac{N_{\text{train}}}{B} \times \text{num\_epochs}
	\end{equation}
	where:
	$N_{\text{train}} = 734002$ -- the number of examples in the training sample, $B = 128$ -- the size of the batch, $\text{num\_epochs} = 20$ -- the number of epochs of training (see Tabs.\,\ref{tab1} and \ref{tab2}).
	The loss is logged at regular intervals to monitor the training process.
	
	\begin{table}[ht]
		\caption{Parameters Setting For Model}
		\begin{center}
			\begin{tabular}{|c|c|}
				\hline
				\textbf{Parameter}&{\textbf{Value}} \\
				\hline
				Learning Rate&  0.001  \\
				\hline
				Optimizer&  Adam  \\
				\hline
				Batch Size&  128 \\
				\hline
				Epochs&  20 \\
				\hline
				Device&  CPU \\
				\hline
			\end{tabular}
			\label{tab2}
		\end{center}
	\end{table}
	
	All baseline machine learning models were implemented using Scikit-learn with default parameters. This choice ensures reproducibility and avoids the potential bias of manual tuning. The KAN model was trained using parameters derived from empirical validation, with no extensive hyperparameter optimization, as the focus was on evaluating baseline feasibility and interpretability.
	
	\section{Experimental Results} \label{sec:experiments}

	This section presents the results of the classification models evaluated for IoT intrusion detection. The proposed KAN model was implemented using Python 3.12 and the PyTorch 12.2 library, running on a Samsung laptop equipped with an Intel Core i7 processor (1.80 GHz $\times$ 8) and Mesa Intel UHD Graphics 620 (KBL GT2). The system runs Linux Mint 21.3 Virginia 64-bit, with Linux version 5.15.0-126-generic. All experiments were carried out with the parameter settings summarized in Table \ref{tab2}, ensuring consistency and reproducibility between models.
	
	Table \ref{tab3} compares the performance of several machine learning models for intrusion detection in IoT security systems. Two experiments were conducted: one using a complete set of features (46 features), and the other using the top 10 features selected based on the importance of the Random Forest feature. The models were evaluated in terms of precision, recall, F1-score, training time, and prediction time. The results of the application of the following methods are given: Logistic Regression (LG), Random Forest (RF), Decision Tree (DT), K-Nearest Neighbors (KNN), Gradient Boosting (GB), XGBoost (XGB), Naive Bayes (NB), Multi-Layer Perceptron (MLP), AdaBoost (AB). For all models implemented using the \textit{sklearn} library, the default parameters were utilized.
	
	\begin{table}[ht]
		\centering
		\caption{Performance Comparison of Different Machine Learning Models (With Top 10 (T10) Features vs. Full Features (Full))}
		\begin{tabular}{||p{2cm}|c|c|c|c|c|c|c|c|}
			\hline
			\multirow{2}{*}{\textbf{Model}} & \multicolumn{2}{c|}{\textbf{Precision}} & \multicolumn{2}{c|}{\textbf{Recall}} &\multicolumn{2}{c|}{\textbf{F1-score}} & \multirow{2}{*}{\begin{tabular}[c]{@{}l@{}}\textbf{Training} \\ \textbf{Time} (s)\end{tabular}} & \multirow{2}{*}{\begin{tabular}[c]{@{}l@{}}\textbf{Prediction} \\ \textbf{Time} (s)\end{tabular}} \\ \cline{2-7}
			& 0 & 1 &  0 & 1 & 0 & 1 &  &  \\ \hline \hline
			LR (Full) & 0.99 & 0.78 & 1.00 & 0.73 & 0.99 & 0.75 & 60.6060 & 0.0155 \\ \hline
			LR (T10) & 0.99 & 0.66 & 0.99 & 0.57 & 0.99 & 0.61 & 4.2741 & 0.0036 \\ \hline \hline
			RF (Full) & 1.00 & 0.93 & 1.00 & 0.94 & 1.00 & 0.94 & 24.2105 & 0.2301 \\ \hline
			RF (T10) & 1.00 & 0.93 & 1.00 & 0.94 & 1.00 & 0.94 & 83.1187 & 0.6880 \\ \hline\hline
			DT (Full) & 1.00 & 0.92 & 1.00 & 0.92 & 1.00 & 0.92 & 3.6391 & 0.0122 \\ \hline
			DT (T10) & 1.00 & 0.91 & 1.00 & 0.92 & 1.00 & 0.92 & 4.9156 & 0.0123 \\ \hline\hline
			kNN (Full)& 1.00 & 0.79 & 0.99 & 0.88 & 1.00 & 0.83 & 0.5305 & 246.5834 \\ \hline
			kNN (T10)& 1.00 & 0.79 & 0.99 & 0.88 & 1.00 & 0.83 & 0.0411& 204.9660
			\\ \hline\hline
			GB (Full)& 1.00 & 0.92 & 1.00 & 0.94 & 1.00 & 0.93 & 289.5622 & 0.2128 \\ \hline
			GB (T10)& 1.00 & 0.92 & 1.00 & 0.94 & 1.00 & 0.93 & 352.8267 & 0.1837 \\ \hline \hline
			XGB (Full) & 1.00 & 0.90 & 1.00 & 0.95 & 1.00 & 0.92 & 3.1733 & 0.0561 \\ \hline
			XGB (T10)& 1.00 & 0.91 & 1.00 & 0.95 & 1.00 & 0.93 & 3.9836 & 0.0729 \\ \hline\hline
			NB (Full)& 1.00 & 0.05 & 0.52 & 1.00 & 0.69 & 0.09 & 1.3125 & 0.1552 \\ \hline
			NB (T10) & 1.00 & 0.05 & 0.53 & 1.00 & 0.69 & 0.09 & 0.4370 & 0.0799 \\ \hline\hline
			MLP (Full) & 1.00 & 0.84 & 1.00 & 0.92 & 1.00 & 0.88 & 793.9634 & 2.1384 \\ \hline
			MLP (T10) & 1.00 & 0.83 & 1.00 & 0.92 & 1.00 & 0.87 & 427.7701 & 1.1884 \\ \hline\hline
			AB (Full) & 1.00 & 0.90 & 1.00 & 0.93 & 1.00 & 0.91 & 72.0482 & 0.9977 \\ \hline
			AB (T10) & 1.00 & 0.90 & 1.00 & 0.93 & 1.00 & 0.91 &72.8655 & 0.9829\\ \hline\hline
		\end{tabular}
		\label{tab3}
	\end{table}
	
	Selecting the top 10 features based on the importance of the random forest feature significantly improved the computational efficiency of most models. For example, the training time for logistic regression (LR) was reduced by 93\% and the prediction time by 77\%. However, a decrease in the recall for malicious traffic (from 0.73 to 0.57) suggests potential information loss.
	
	Based on the analyses, the best overall models for the task of intrusion detection in IoT networks are Random Forest (RF)  and XGBoost (XGB). These models offer a strong balance between precision, recall, F1 score, and training time, making them ideal for detecting benign and malicious traffic effectively. Although feature selection significantly improved efficiency for models like Logistic Regression and MLP, but at the cost of reduced recall to detect malicious traffic in LR.
	
	Both KNN and Naive Bayes had significant drawbacks, with KNN's high prediction time and Naive Bayes poor recall limiting their practicality for IoT intrusion detection tasks, even after feature selection. Decision trees were shown to be highly efficient across both experiments, with stable performance and minimal computational costs, making them a viable option for simple classification tasks in IoT systems.
	
	The goal of the next experiment was to evaluate the performance of a KAN architecture for intrusion detection in IoT systems. The KAN model was tested on a selected set of features from the dataset, with a focus on analyzing accuracy, loss reduction, and computational efficiency (training and prediction time). The results of the network construction are shown in Table \ref{tab4}. The results showed a performance comparison between the KAN model and various traditional machine learning models in terms of recall, F1-score, training time and prediction time, both using the full set of features (Full) and a selection of the 10 most relevant features (T10).
	
	\begin{table}[ht]
		\centering
				\caption{Training and Evaluation Results for KAN Model (With Top 10 Features (T10) vs. Full Features (Full))}
		\begin{tabular}{|p{2.3cm}|c|c|c|c|c|c|c|c|}
			\hline
			\multirow{2}{*}{\textbf{Model}} & \multicolumn{2}{c|}{\textbf{Precision}} & \multicolumn{2}{c|}{\textbf{Recall}} &\multicolumn{2}{c|}{\textbf{F1-score}} & \multirow{2}{*}{\begin{tabular}[c]{@{}l@{}}\textbf{Training} \\ \textbf{Time} (s)\end{tabular}} & \multirow{2}{*}{\begin{tabular}[c]{@{}l@{}}\textbf{Prediction} \\ \textbf{Time} (s)\end{tabular}} \\ \cline{2-7}
			& 0 & 1 &  0 & 1 & 0 & 1 &  &\\ \hline \hline
			KAN (Full) &1.00 & 0.70 & 0.99 & 1.00 & 0.99& 0.82 & 26720.4515 & 3.2056 \\ \hline
			KAN (T10) & 0.99 & 0.61 & 0.99 & 0.48 & 0.99& 0.54& 864.4360 & 0.7326 \\ \hline \hline
		\end{tabular}
		\label{tab4}
	\end{table}

	The KAN model demonstrated strong learning capacity, reducing the loss from 0.0637 to 0.0239 over 20 training steps. For malicious traffic (class 0), KAN achieved near-perfect precision (1.00) and recall (0.99) with full features, yielding an F1-score of 0.99. For benign traffic (class 1), the precision dropped to 0.70, but the recall remained perfect (1.00), resulting in an F1-score of 0.82. With the top 10 characteristics (T10), the precision for class 0 remained high (0.99), but the recall dropped to 0.48 (F1: 0.54), highlighting a critical trade-off between efficiency and detection reliability.
	
	Training times for KAN were substantial, requiring 7 hours for full features and 14 minutes for T10, far exceeding tree-based models like Random Forest (24 seconds) or XGBoost. However, KAN’s prediction times remained practical for deployment at 3.2 seconds (full dataset) and 0.73 seconds (T10). While RF and XGBoost outperformed KAN in class 0 F1-scores (0.94 and 0.93 vs. 0.82), KAN surpassed simpler models like Logistic Regression (class 0 F1: 0.61) and K-Nearest Neighbors (class 0 F1: 0.83), balancing performance with interpretability.
	
	Feature selection reduced computational costs but degraded detection metrics. For Logistic Regression, recall for malicious traffic dropped sharply from 0.73 to 0.57, lowering its F1-score from 0.75 to 0.61. Similarly, KAN's class 0 F1 score decreased to 0.54 (from 0.82) with T10, although it retained high precision (0.99). This underscores the challenge of maintaining detection efficacy when prioritizing computational efficiency.
	
	A key strength of KAN lies in its interpretability. Unlike black-box models such as RF or XGBoost, KAN’s learnable spline activations generate symbolic formulas, clarifying feature interactions critical for diagnosing false positives in malicious traffic predictions. This transparency enables precise tuning of security systems, even when modeling complex non-linear relationships.
	
\begin{figure}[tb]
		\centering
		\includegraphics[width=0.7\linewidth]{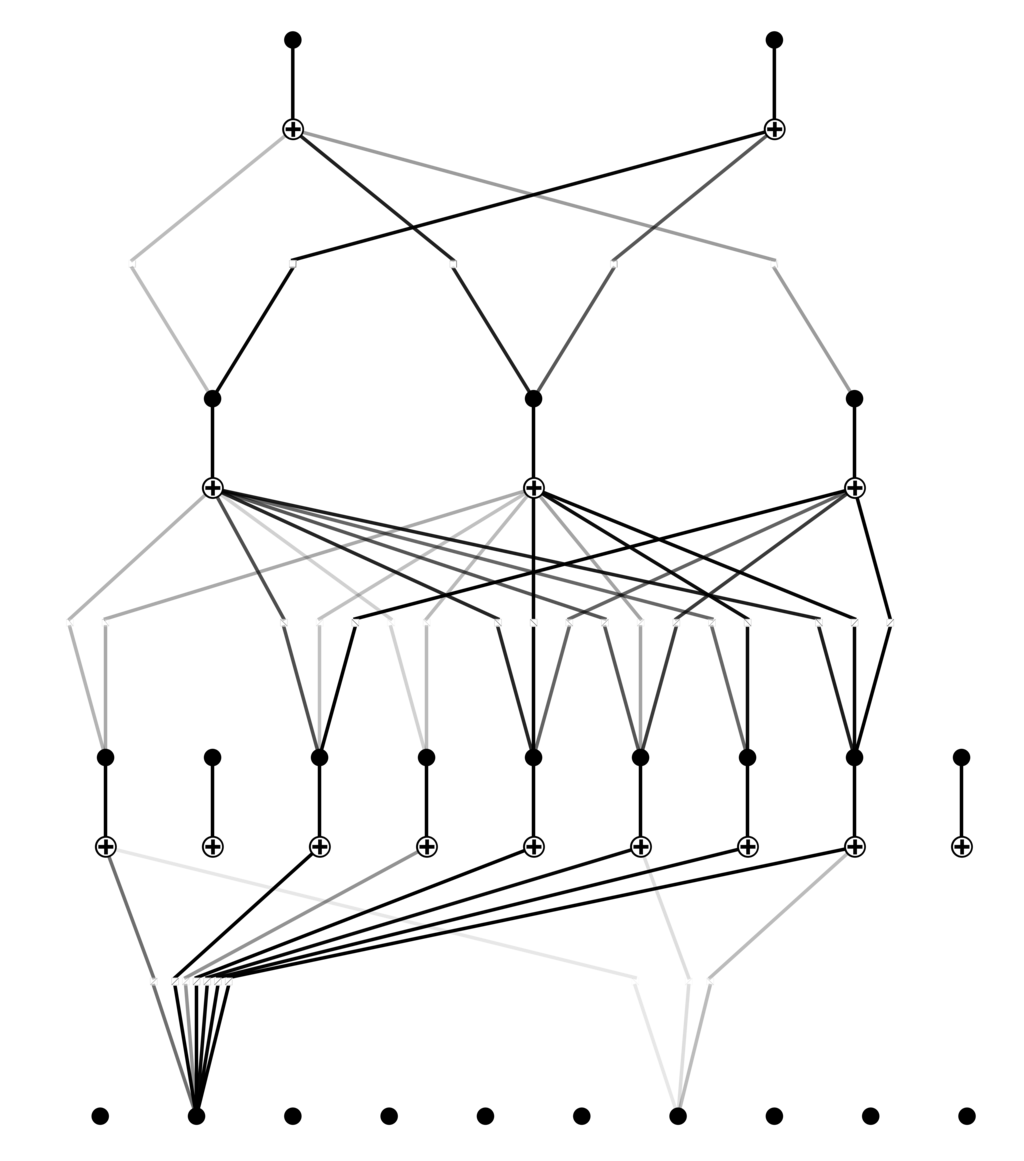}
		\caption{Representation of the KAN architecture for Model (Top 10).}
		\label{fig5}
	\end{figure}

\begin{figure}[tb]
		\centering
		\includegraphics[width=0.7\linewidth]{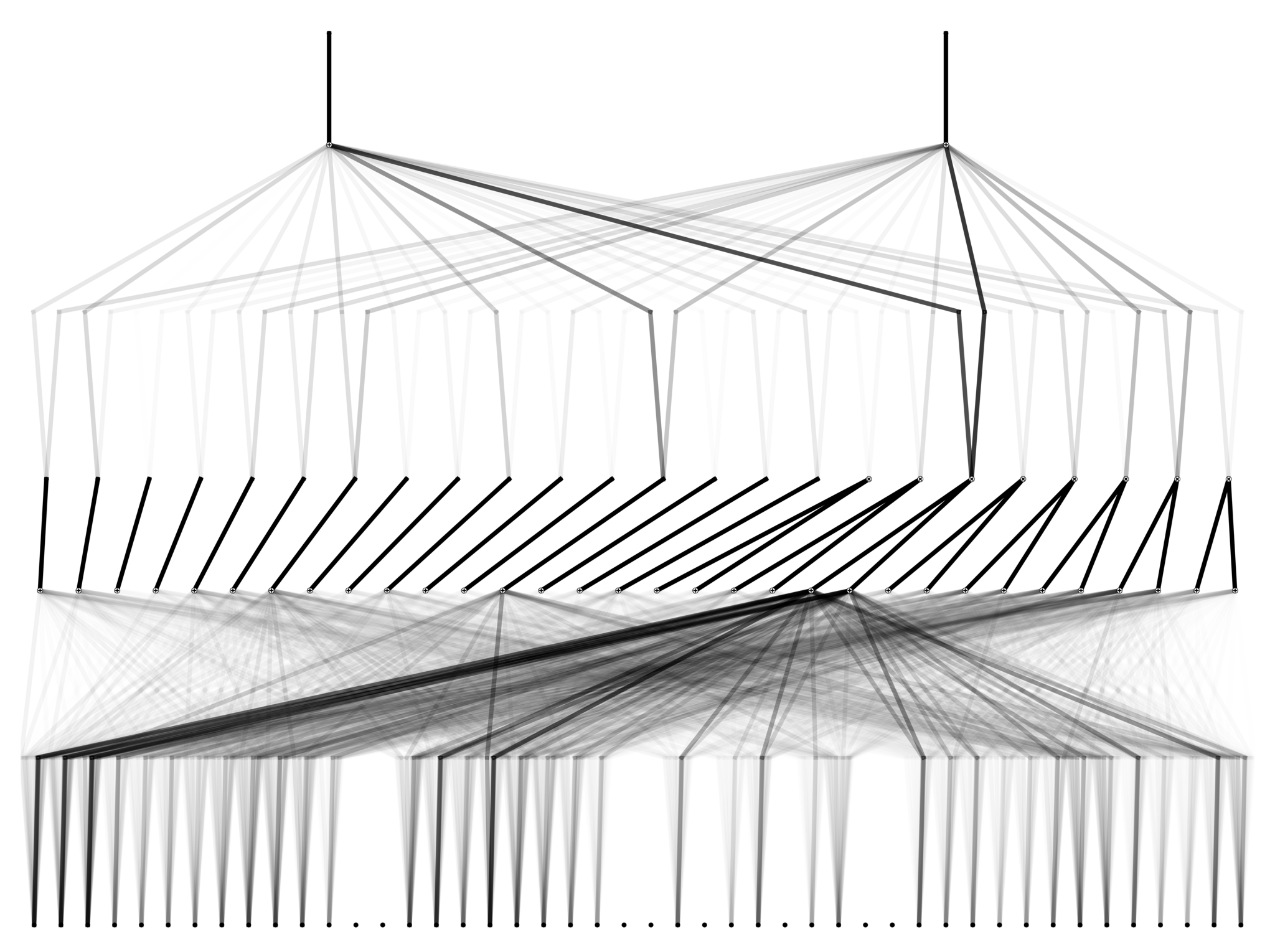}
		\caption{Representation of the KAN architecture for Model (Full).}
		\label{fig6}
	\end{figure}

	A visual representation of the KAN architecture is shown in Fig.\,\ref{fig5} e \ref{fig6}. 
	The lower row of nodes represents the input features of the data set according to the selected importance of the feature applied (see Sec.\,\ref{sec:method_3}). The thickness and opacity of the links represent the strength or importance of these learned functions. The top layer represents the final classification result, whether the traffic is classified as benign or malicious.
	
	Eq.\,\ref{eq4} represents a symbolic formula generated by the KAN model for the complete dataset. The equations generated by the KAN model demonstrate the ability to capture complex and non-linear relationships between variables, combining linear, non-linear and trigonometric terms. This makes it highly effective in detecting complex patterns in IoT environments, where data can exhibit periodic variations. However, the model also presents challenges in terms of computational cost, particularly due to the complexity of the equations, which affects efficiency in large-scale or real-time scenarios. Generalization is well balanced, but efficiency improvements are needed to make it more competitive with traditional ML models.
	
	{\small
		\begin{equation} \label{eq4}
			\begin{aligned}
				f(x_1, \dots, x_{46}) &= -0.1319 A \cdot B + 5.9730\\
				&\quad  - 0.1890 A \cdot B - 3.5679,
			\end{aligned}
	\end{equation}}
	where
	{\small
		\begin{equation}\label{eq4_1}
			\begin{aligned}
				A &= 0.3865 x_1 - 0.2209 x_{10} + 0.0810 x_{11} + 0.1644 x_{15} \\
				&\quad + 0.2828 x_{16} + 0.1019 x_{17} + 0.1527 x_{18} - 0.1815 x_{19} \\
				&\quad - 0.4414 x_2 + 0.1003 x_{20} - 0.1394 x_{21} + 0.1011 x_{22} \\
				&\quad - 0.1073 x_{25}  - 0.1079 x_{28} - 0.2856 x_{34} - 0.1251 x_{35}\\
				&\quad  + 0.1145 x_{36} - 0.1775 x_{39} + 0.0795 x_5 + 0.1521 x_9 \\
				&\quad + 1.0500 \sin(0.5490 x_3 + 8.4495)\\
				&\quad  + 0.3195 \sin(0.4637 x_{41} - 0.8215)\\
				&\quad  + 0.2710 \sin(0.4930 x_{45} - 1.2037) - 0.2464,
			\end{aligned}
	\end{equation}}
	and
	{\small
		\begin{equation}\label{eq4_2}
			\begin{aligned}
				B &= 0.5147 x_1 - 0.0708 x_{11} - 0.1260 x_{12} - 0.0617 x_{17}\\
				&\quad  + 0.4938 x_{18} - 0.2913 x_2 + 0.0651 x_{20} - 0.2176 x_{25}\\
				&\quad  + 0.0609 x_{27} - 0.2009 x_{28}  + 0.0801 x_{30} - 0.1499 x_{34}\\
				&\quad  + 0.1205 x_{36} - 0.0881 x_{38} + 0.2016 x_4  + 0.0629 x_{43}\\
				&\quad  + 0.0723 x_5 - 0.1280 x_6 + 0.0450 x_7 + 0.1113 x_8 \\
				&\quad - 0.2193 \sin(0.4226 x_{41} + 1.3504)\\
				&\quad  + 0.8037 \sin(0.2836 x_{46} - 7.4806) \\
				&\quad + 1.3886 \cos(0.3002 x_{16} - 2.5864)\\
				&\quad  + 0.8571 \cos(0.6163 x_3 - 5.6287) \\
				&\quad - 6.6183 \cos(0.0921 x_{40} + 6.4281) \\
				&\quad + 0.6129 \cos(0.6235 x_{45} - 9.1999) + 8.1104.
			\end{aligned}
	\end{equation}}

	For real-time IoT deployments, Random Forest and XGBoost are good choices due to their rapid training times and superior class 0 F1-scores (0.94). However, KAN is ideal for scenarios that prioritize model transparency or require granular analysis of attack patterns, despite its computational overhead. Future work could explore hybrid architectures that integrate KANs with tree-based feature selection to optimize both efficiency and interpretability, bridging the gap between real-time performance and actionable insights in IoT security systems.

	\section{Discussion} \label{sec:discussion}
	This study evaluated the performance of a Kolmogorov-Arnold network architecture compared to traditional machine learning models for intrusion detection in Internet of Things systems. The results provided insight into the accuracy, training time, and prediction time of different models when applied to a comprehensive dataset of IoT network traffic, with additional analysis conducted using a reduced feature set based on feature importance.
	
	The results also highlight the potential capabilities of KANs over traditional neural network architectures. By using learnable activation functions at edges, KANs can capture complex, non-linear relationships more effectively than models based on fixed activation functions such as MLPs. This flexibility and adaptability make KANs a promising alternative for security applications in the IoT, where network traffic can exhibit a wide range of behaviors that need to be interpreted dynamically.
	
	The trained KAN model can be used for intrusion detection in real-time IoT systems, providing a robust and scalable solution for detecting both known and unknown threats. Future work could focus on refining the model to further reduce training time through optimization techniques and the use of GPU acceleration. In addition, exploring the application of federated learning could enable collaborative model training across distributed IoT networks, improving privacy while maintaining high model performance.
	
	Random Forest and XGBoost remain the top choices for IoT intrusion detection, primarily due to their high accuracy and fast computation times. These models are highly scalable, making them ideal for real-time IoT security environments where resource constraints and real-time response are critical.
	
	Although KAN is highly accurate and capable of approximating complex functions, it is better suited for applications where interpretability and function approximation are priorities. Its longer training time currently limits its suitability for large-scale or real-time IoT applications, although its ability to model complex data interactions could be highly beneficial in scenarios where more intricate relationships in the data need to be explored.
	
	This work demonstrates that Kolmogorov–Arnold Networks (KANs) are a viable approach for intrusion detection systems (IDS) due to their inherent interpretability. Unlike traditional models that require external XAI techniques (e.g., SHAP, LIME), KANs generate symbolic formulas directly from the model, enabling more transparent and auditable decision-making~---~a critical requirement in regulated or mission-critical IoT environments.
	
	Although training time remains a limitation, KANs are particularly valuable in scenarios where interpretability is as important as accuracy. Their potential is especially relevant in federated learning or edge-based IoT systems, where privacy, decentralization, and explainability must coexist. Future research may explore hardware acceleration (e.g, the use of Graphics Processing Unit~---~GPUs) and hybrid pipelines that combine symbolic reasoning with efficient feature selection, helping bridge the gap between theoretical innovation and real-world applicability.
	
	This work contributes to the growing body of research aimed at improving IoT security through machine learning, particularly in environments where resource constraints and real-time requirements are prominent.
	
	This study shows that KANs are accurate and interpretable, which is vital for transparent, auditable detection processes. While models like Random Forest and XGBoost are excellent, they lack the explainability of KANs. This capacity for transparent decision-making is valuable for scenarios like regulatory compliance, forensic analysis, or critical infrastructure protection, where understanding the rationale behind alerts is as important as detection accuracy itself.
	
	\section{Conclusion}\label{sec:conclusion}
	
	The study demonstrates that Kolmogorov-Arnold Networks (KANs) are highly effective at capturing complex, non-linear relationships in IoT environments, significantly outperforming traditional machine learning models. The research emphasizes the crucial importance of optimized feature selection, which not only improves model performance by reducing the number of variables but also minimizes training time and computational overhead, thereby facilitating real-time application in resource-constrained IoT systems.
	
	The integration of KANs with learnable activation functions represents a significant advancement in the field of IoT security frameworks. This integration provides a robust solution that improves both the accuracy and interpretability of network traffic classification, which is essential for the protection of sensitive data from evolving cyber threats. As future work, optimizing KANs via hardware acceleration (GPUs) could bridge the training efficiency gap, making them viable for large-scale IoT deployments.
	
	In future applications, KANs could be especially beneficial in domains where the ability to generate interpretable detection rules is critical. Hybrid approaches combining interpretable symbolic layers with faster detection backends could allow balancing detection performance with transparency.
	
	\section*{Acknowledgement}
	
	This study was financed in part by the Coordena\c{c}\~{a}o de Aperfei\c{c}oamento de Pessoal de N\'{\i}vel Superior - Brasil (CAPES) - Finance Code 001.
	
%

\end{document}